\documentclass{egpubl}
\usepackage{sca2020}

\usepackage[T1]{fontenc}
\usepackage{dfadobe}  

\usepackage{cite}  
\BibtexOrBiblatex
\electronicVersion
\PrintedOrElectronic
\ifpdf \usepackage[pdftex]{graphicx} \pdfcompresslevel=9
\else \usepackage[dvips]{graphicx} \fi

\usepackage{egweblnk} 

\title%
      {Learning to Generate Diverse Dance Motions with Transformer}

\author[Jiaman Li, Yihang Yin, Hang Chu, Yi Zhou, Tingwu Wang, Sanja Fidler \& Hao Li]
{\parbox{\textwidth}{\centering Jiaman Li$^{1,2}$
Yihang Yin$^{3}$
Hang Chu$^{4,5}$
Yi Zhou$^{1}$
Tingwu Wang$^{4,5}$
Sanja Fidler$^{4,5}$
        and Hao Li$^{6}$
        }
        \\
{\parbox{\textwidth}{\centering $^1$University of Southern California
         $^2$USC Institute for Creative Technologies 
         $^3$Beihang University \\
         $^4$University of Toronto 
         $^5$Vector Institute 
         $^6$Pinscreen
      }
}
}

\usepackage{pifont}
\usepackage{hhline}
\usepackage{multirow}
\usepackage{amsmath}
\usepackage{amssymb}
\usepackage{xcolor}

\newcommand{\nothing}[1]{}
\definecolor{DeltaColor}{rgb}{0.039,0.73,0.71}
\definecolor{SigmaColor}{rgb}{0.98,0.45,0.0}
\definecolor{AlphaColor}{rgb}{0,0,0.8}
\definecolor{BetaColor}{rgb}{0.8,0,0.8}
\definecolor{GammaColor}{rgb}{0.514,0.34,0.224}
\definecolor{EpsilonColor}{rgb}{0.353,0.725,0.906}
\definecolor{YajieColor}{rgb}{0.7,0.5,1.0}
\newcommand{\yajie}[1]{{\color{YajieColor} Yajie: #1 $\qed$}}

\begin{document}
\newcommand{\cmark}{\ding{51}}%
\newcommand{\xmark}{\ding{55}}%
\teaser{
\vspace{-8mm}
 \includegraphics[width=1\linewidth]{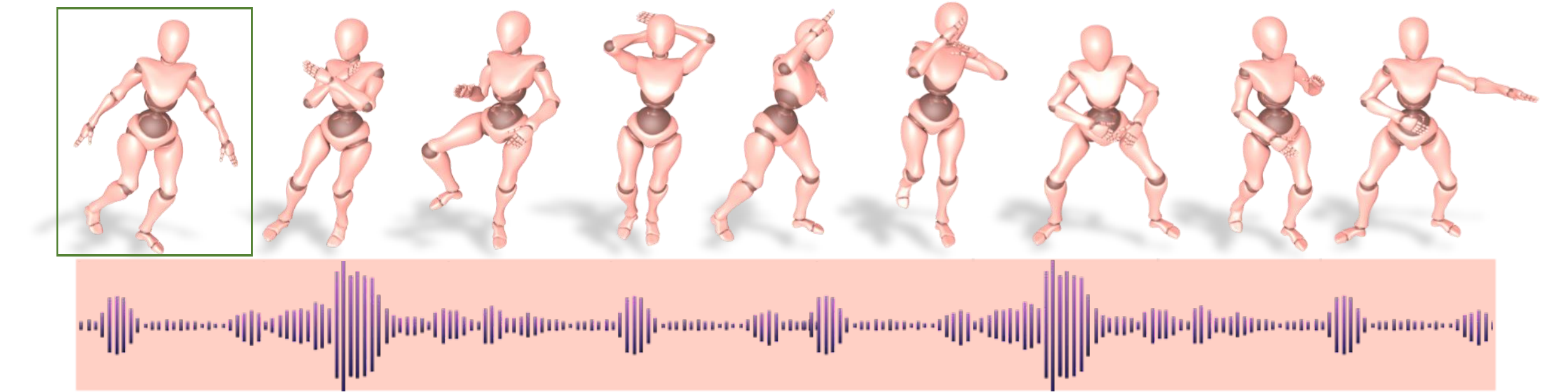}
 \centering
  \caption{Given novel music (shown in the second row), our model generates diverse dance motions following beats (shown in the first row). The green box shows the initial pose for dance motion synthesis.}
\label{fig:teaser}
}

\maketitle

\begin{abstract}
\nothing{
\yajie{
* What s the problem?
\begin{enumerate}
    \item The current industrial pipeline of dance motion synthesis is inefficient and at a high cost.
\end{enumerate}
* What do we introduce?
\begin{enumerate}
    \item We introduce a framework of automatically dance motion synthesis with large diversity and complexity in motion.
\end{enumerate}
* How does it roughly work? (main differentiator to existing methods? What’s unique?)
\begin{enumerate}
    \item The diversity in our database compared to small scale mocap data
    \item network unique merits
     \item introduced a benchmark to evaluate/select the best pick 
\end{enumerate}

* What can we do now? 

\begin{enumerate}
    \item We can generate novel and complex dance motion following the music
    \item we can score the synthesized sequences.
\end{enumerate}

* What do we show?
\begin{enumerate}
    \item briefly describe your advances in the results

\end{enumerate}

* What’s the impact?
\begin{enumerate}
    \item show how vast online data can be used to bootstrap the lack of mocap data.
    \item greatly reduce the production cycle of dance moves in a professional pipeline by providing initial and inspiration
    \item provide an easily accessible solution for low-cost motion synthesis-related application 

\end{enumerate}

}
}
With the ongoing pandemic, virtual concerts and live events using digitized performances of musicians are getting traction on massive multiplayer online worlds. However, well choreographed dance movements are extremely complex to animate and would involve an expensive and tedious production process. In addition to the use of complex motion capture systems, it typically requires a collaborative effort between animators, dancers, and choreographers. We introduce a complete system for dance motion synthesis, which can generate complex and highly diverse dance sequences given an input music sequence. As motion capture data is limited for the range of dance motions and styles, we introduce a massive dance motion data set that is created from YouTube videos. We also present a novel two-stream motion transformer generative model, which can generate motion sequences with high flexibility. We also introduce new evaluation metrics for the quality of synthesized dance motions, and demonstrate that our system can outperform state-of-the-art methods. Our system provides high-quality animations suitable for large crowds for virtual concerts and can also be used as reference for professional animation pipelines. Most importantly, we show that vast online videos can be effective in training dance motion models. 


\end{abstract}  
\section{Introduction}

Due to the ongoing COVID-19 pandemic, an entire global live events industry is being shut down. Interactive Vtuber performances and virtual music concerts that take place in online gaming platforms are becoming increasingly popular. Examples include the widely popular holographic and VR concerts with Hatsune Miku, or the virtual rap event in Fortnite, performed by a digital avatar of Travis Scott, hosting tens of millions of viewers in real-time in a massive multiplayer online setting. The ability to generate complex and believable dance motions through music alone could impact a broad range of interactive and immersive applications in entertainment. In existing production settings, dance animations are typically generated through carefully choreographed performances and they are often captured using complex motion capture systems. Intensive manual labor is generally required for data clean up and animation refinement, which results in long and expensive production cycles.

More recently, data-driven motion synthesis methods~\cite{holden2016deep,li2017auto} have been introduced to scale the production of animation without the need of actual performers. However, most of these techniques are based on a very limited diversity of motion, \textit{e.g.} the popular CMU mocap dataset~\cite{cmumocap} in~\cite{li2017auto} containing only two kinds of dances with a duration of less than one hour. These sequences are often very short and difficult to expand. Furthermore, the movements are often monotonous with repetitive patterns and contain little variation and diversity. Therefore, models trained on these mocap data show limited generalization capabilities for realistic dance motions.

With recent advances in learning-based motion synthesis, several deep learning-based methods have been introduced~\cite{holden2016deep,li2017auto}. These techniques formulate the problem as a long-term prediction task~\cite{li2017auto} or as an audio-to-pose sequence translation problem~\cite{shlizerman2018audio, alemi2017groovenet, augello2017creative}. These regression-based methods fail to predict highly diverse and complex motions even using advanced training strategies via RNN models~\cite{li2017auto} as they are deterministic. In particular, these models are designed to predict the successors given the current status instead of generating a novel motion according to a database distribution. These methods are not suitable for music-oriented dance motion synthesis which usually expects more than one possible motions given the same music. We further note that there is no benchmark or evaluation metric other than visual inspection for all the above methods.

In this work, we introduce a novel system that can synthesize diverse dance motions by learning from a large-scale dataset with a comprehensive set of highly diverse dance movements. To achieve this goal, we face three major challenges: (1) it is difficult to physically collect a large-scale dance motion dataset with sufficient diversity. Although motion/performance capture methods can provide high-precision data, they require dedicated devices, professional performers, and a tedious clean-up process; (2) existing regression-based models cannot handle the diversity of dance movements; (3) long-term temporal consistency and motion coherency have to be preserved in the generated sequence.

To address the above issues, (1) we take advantage that a large number of dance videos are available online. We download thousands of dance videos, and use cutting edge techniques for 2D pose detection, tracking, and 3D pose estimation to recover the dance sequence. Our large-scale dance motion dataset consists of 50 hours of s synchronized music and dance pose sequences.
(2) Using this dataset, we propose a conditional auto-regressive generative model to learn the motion distribution, along with Transformers~\cite{vaswani2017attention} as the main architecture for capturing extended time dependency. We formulate the output in each timestep as a categorical distribution using the discrete pose representations inspired by the success of discrete audio representation used in WaveNet architecture~\cite{oord2016wavenet}. The discrete pose representation enables us to model the next step's pose distribution and sample diverse poses at inference. Our model not only outperforms previous models for important evaluation metrics but also enables generating diverse dances with new music, demonstrating better modeling and generalization capabilities.
(3) Besides, we propose several evaluation metrics from different perspectives to better judge whether the motion synthesis is satisfactory or not. We first use a Bullet-based~\cite{coumans2013bullet} virtual humanoid to evaluate the feasibility of generated pose sequences. Then, inspired by the motion-beat analysis approach in~\cite{kim2003rhythmic}, we introduce an effective and automatic metric for evaluating whether the dance movements follow the beat properly. In addition to physical plausibility and beat consistency, we also provide a metric for dance variation to measure the diversity in the synthesized results.

By testing our model using different settings and comparing it with two main baseline techniques, including acLSTM~\cite{li2017auto} and ChorRNN~\cite{crnkovic2016generative2} in a non-audio setting, we show that our real-time method can generate more diverse and realistic dance motions than existing techniques. We also show that when compared with LSTM architectures, our Transformer model is also more efficient to train. Finally, we demonstrate the effectiveness of our model and proposed evaluation metrics using a perceptual study. Our main contributions include:
\begin{enumerate}
    \item An end-to-end real-time system for dance motion synthesis that uses highly complex and diverse motion data obtained from Internet videos. We also introduce an efficient and scalable data collection pipeline.
    \item A novel two-stream motion transformer model with discrete pose representation to model the motion distribution and to capture long-term dependencies, which can be conditioned on music for diverse dance motion synthesis. 
    \item Several effective evaluation metrics to assess the quality of synthesized dance motions.
\end{enumerate}

\vspace{-2mm}
\section{Related Work}
Dance motion synthesis is a highly interdisciplinary problem and we review the most relevant work here.

\vspace{-2mm}
\subsection{Motion Synthesis}
Motion synthesis has been an actively studied problem in both, the computer graphics and the computer vision communities. 
Typical methods rely on representation learning techniques such as auto-encoders, to embed motions into a low-dimensional space.
Convolutional auto-encoders have been used to learn valid motion representation termed as the motion manifolds~\cite{holden2015learning}, such that corrupted motion or missing-marker data can be easily recovered. 
High-level, user-friendly parameters for motion control have also been  explored~\cite{holden2016deep}. In particular, they first specify character trajectory and foot contact information, then conditionally synthesize the human motion. 
A language-guided motion generation framework~\cite{lin:vigil18} has been proposed to generate realistic motions from natural language descriptions. 

Motion synthesis can be also viewed as a long-term human motion prediction. A number of works use recurrent neural networks to address this problem. In the work of Martinez~et~al.~\cite{martinez2017human}, practical strategies including adding residual blocks and introducing sampling during training to improve RNN learning.
Auto-conditioned RNN~\cite{li2017auto} takes both ground truth and model prediction as input with a specific alternating interval to train the sequential model, showing the potential of generating motion sequences with long, or even unlimited future duration.   
QuaterNet~\cite{pavllo:quaternet:2018} conducts extensive experiments to demonstrate the effectiveness of quaternion representation.
Dance generation can be regarded as a special case of motion synthesis, while dance movements are more complex compared to usual motions such as walking. It is also important to ensure the coordination between music and motions. 
\vspace{-4mm}
\subsection{Choreography Learning}
The creative process of choreography requires rich dancing experiences, which has inspired the development of data-driven approaches. A common approach of predicting dance movements from audio is to formulate the task as a translation problem. In particular, this approach enforces a one-to-one mapping from music to dance, which does not generalize well beyond training songs.
GrooveNet~\cite{alemi2017groovenet} has been the first work to exploit RNNs to model the correspondence between audio features and motion capture data. 
Based on music content analysis with motion connectivity constraints, a probabilistic framework~\cite{fukayama2015music} has been proposed. It mainly focuses on exploring and identifying the major contributing choreographic factors. 
Moreover, auto-encoder~\cite{augello2017creative} has been exploited for mapping music features to a latent space, and to generate dance pose sequences. Essentially, this formulation still falls into the audio to dance mapping category.
Similar to previous work in choreography, Audio to Body Dynamics~\cite{shlizerman2018audio} uses time-delayed RNNs to learn a mapping from audio to hand key points. The common drawbacks of these works lies in the difficulty of generating diverse pose sequences given audio information only. We argue that dance movements are more complex, hence less suitable to such deterministic formulation. This is because the same music can induce various kinds of dance motions, and the aesthetics of dance is strongly tied to the diversity of motions. 
Other work solve the task from a generative modeling perspective, ChorRNN~\cite{crnkovic2016generative}, which introduces mixture-density RNNs to generate pose sequences and is particularly promising. Nevertheless, their approach is motion-only and does not take music information into account.  
In this paper, we treat dance motion synthesis as a conditional generative task. This allows us to not only promote the diversity of generation, but also take music as conditions to ensure the consistency with music features.


\begin{figure*}[h]
\begin{center}
\includegraphics[width=\textwidth]{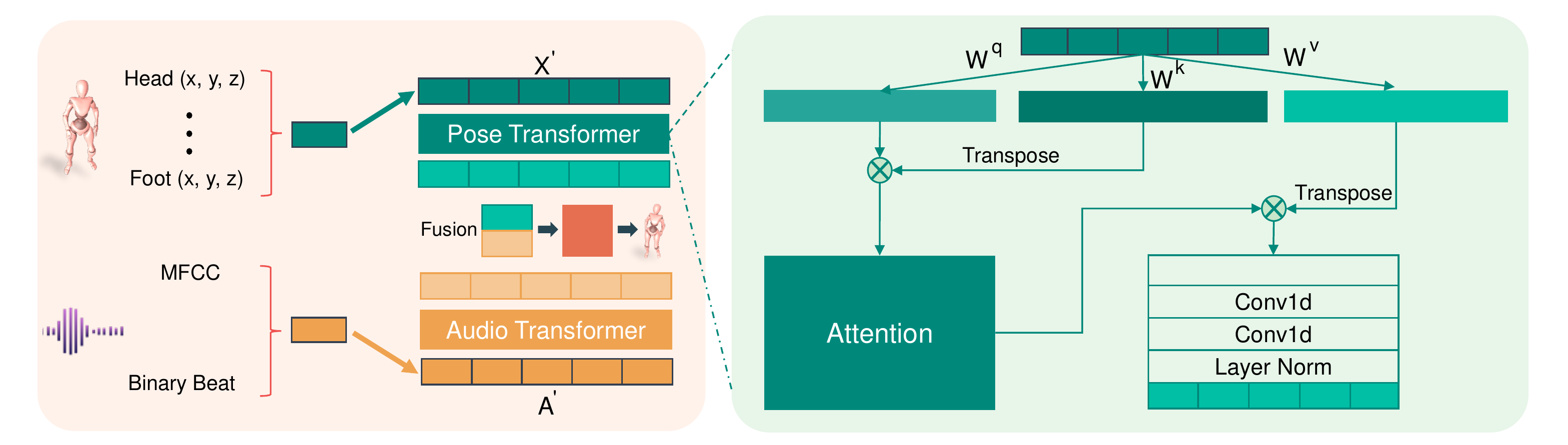}
\end{center}
\vspace{-2mm}
\caption{Overview of our TSMT model. Left shows an example time step of pose and audio embedding. Middle shows our Two-Stream Motion Transformer including a pose transformer and an audio transformer. Each time step is followed by a late fusion module between two streams, which predicts the pose in the next time step. Right shows a detailed sub-layer composition inside a transformer block.
}
\label{model_pipeline}
\vspace{-1mm}
\end{figure*}

\vspace{-4mm}
\section{Methods}

In this section, we present the Two-Stream Motion Transformer (TSMT) model. It processes the pose sequence and music context separately and then fuses the two streams together to predict the next motion. 

\vspace{-3mm}
\subsection{Problem Formulation}

Inspired by the recent success of the auto-regressive model~\cite{oord2016wavenet}, we formulate our problem as an auto-regressive generative model conditioned on both music and past motions for synthesizing realistic and self-coherent dance motions. We denote the sequence of 3D dance motions as $\mathbf{X}$=\{$\mathbf{x}_1, ..., \mathbf{x}_T$\}, and the sequence of audio features as $\mathbf{A}$=\{$\mathbf{\alpha}_1, ..., \mathbf{\alpha}_T$\}. $T$ is the number of frames in the sequence. 
For each time step, $\mathbf{x}_t$$\in$$\mathbb{R}^{3n}$ and $\mathbf{\alpha}_t$$\in$$\mathbb{R}^{m}$, where $n$ is the number of body joints and $m$ is the dimension of audio features.
We model the joint conditional probability as
\begin{eqnarray}
p(\mathbf{X}) = \prod_{t=1}^T p(\mathbf{x}_t|\mathbf{\alpha}_t, ..., \mathbf{\alpha}_1, \mathbf{x}_{t-1}, ..., \mathbf{x}_{1}) 
\end{eqnarray}
where the motion stream and the audio stream are both encoded by neural networks.

\vspace{-3mm}
\subsection{Motion and Audio Representation}
\subsubsection{Motion}
Unlike acLSTM~\cite{li2017auto} and ChorRNN~\cite{crnkovic2016generative} which represent poses as deterministic coordinates or the distributions respectively, we represent the continuous value of joint coordinates as discrete categories. For each dimension of the 3D pose $\mathbf{x}_t$, we perform uniform discretization into 300 constant intervals and obtain $3n$ 300-dimensional one-hot vectors. To reduce memory cost, we then transform each one-hot vector into a $D^E$-dimensional embedding vector with a shared learnable matrix of size $[D^E, 300]$. This converts the motion sequence into a tensor of size $[T, 3n, D^E]$.
We merge the latter two axes of motion embedding and input it to a temporal-wise fully connected feed-forward layer to obtain a sequence of vectors with $D^M$ channels.
Following Vaswani~et~al.~\cite{vaswani2017attention}, we also compute a $D^M$-dimensional positional embedding sequence with sine and cosine functions to encode the temporal information. We add the positional embedding sequence to the motion embedding sequence. This forms our final motion representation $\mathbf{X}'$ of size $[T, D^M]$.

\vspace{-3mm}
\subsubsection{Audio}
For the audio data at each time step, we directly use the continuous 13-dimensional MFCC vector concatenated with its temporal derivatives into a 26-dimensional feature vector. We embed the binary one-hot beat signal into a 30-dimensional vector, resulting different embedding vectors at beat and non-beat positions. Similar to motion, we feed the audio representation into a 1D convolution and add to the positional embedding. The output is denoted as $\mathbf{A}'$. 

\vspace{-4mm}
\subsection{Two-Stream Motion Transformer (TSMT)}
\subsubsection{Transformer}

We adopt the Transformer~\cite{vaswani2017attention} architecture, harnessing and exploiting its power in modeling sequential data.
Transformer consists of multiple blocks, each block is further composed of multiple heads of self-attention sub-layers as well as position-wise feed-forward sub-layers.
Obtained from the previous step, our input $\mathbf{X}'$ is a matrix of size $[T, D^M]$.
We first transform the input sequence into three matrices, namely keys $\mathbf{K}$=$\mathbf{X}'\mathbf{W}^{K}$, queries $\mathbf{Q}$=$\mathbf{X}'\mathbf{W}^{Q}$, and values $\mathbf{V}$=$\mathbf{X}'\mathbf{W}^{V}$, where $\mathbf{K}$, $\mathbf{Q}$, $\mathbf{V}$ are $[T, D]$ matrices. 
We split each matrix into multiple heads where each head is $[T, D_i]$, with $D$=$\sum_i D_i$. 
For each head, we compute the scaled-dot attentional features as
\begin{eqnarray}
\mathbf{Z}_{i} = \mbox{Softmax}(\frac{\mathbf{Q}_i\mathbf{K}_{i}^{T}}{\sqrt{D_{i}}})\mathbf{V}_i 
\end{eqnarray}
We concatenate all the $\mathbf{Z}_{i}$ to form $\mathbf{Z}$ with size $[T, D]$, then feed it into two position-wise 1D convolution layers followed by layer normalization \cite{vaswani2017attention}.
This forms one Transformer block. We use multiple such blocks with residual connections between the blocks.
We refer readers to \cite{vaswani2017attention} for more details about the Transformer architecture. 


\vspace{-2mm}
\subsubsection{TSMT}
The essence of dance is the manifestation of musicality in physical forms.
Musicality takes form of multiple components such as vocal, bass, snare, keyboard, hi hat, drum, and other sound effects. 
A key element is carefully paying attention to different layers of music. 
Therefore, we use a pose-stream transformer to capture dance history, an audio-stream transformer to extract music context,
and fuse these two streams together to predict the next pose as shown in Figure~\ref{model_pipeline}.
We denote the output of pose transformer as $\mathbf{Z}^X$=\{$\mathbf{z}_1^X, ..., \mathbf{z}_T^X$\}, and the output of audio transformer as $\mathbf{Z}^A$=\{$\mathbf{z}_1^A, ..., \mathbf{z}_T^A$\}. We compute the discrete representation of the pose at next time step from $\mathbf{Z}^X$ and $\mathbf{Z}^A$ and  predict the final pose as,
\begin{equation}
\begin{aligned}
\log p(\mathbf{x}_{t}) & = \mbox{Softmax}(\mathbf{W}^X\mathbf{z}_{t-1}^{X}+\mathbf{W}^A\mathbf{z}_{t}^{A})
\end{aligned}
\end{equation}
During training, we can efficiently compute all the time steps in parallel, with a mask applied to the attention to ensure each step only attend to its past. During inference, we sample from the log-likelihood at the new time step.

\vspace{-4mm}
\subsection{Implementation Details}

The model consists of multiple blocks with multiple layers, each containing the multi-head self-attention and position-wise feed-forward layers as described above. We list important model parameters in Table~\ref{table:detailed_setting}.
To train the models, We use Adam optimizer with mini-batches of size 32. We set the initial learning rate as $10^{-4}$ with $0.3$ decay rate after 200 epochs. The global motion is inferred by the Global Path Predictor in Zhou et al.~\cite{zhou2020generative}.

\begin{table}[t!]
\begin{center}
\setlength{\tabcolsep}{9.5pt}
\footnotesize{
\begin{tabular}{@{}l||ccccccc@{}} 
 \hline
~ & $D^E$  & $D_i$ & $D^M$ & block & layer & head \\
\hline\hline
Pose & 5 & 128 & 256 & 4 & 4 & 4\\
\hline
Audio & - & 32 & 64 & 2 & 2 & 2\\
\hline
\end{tabular}
}
\end{center}
\vspace{-1mm}
\caption{\small Model details. $D^E$ for embedding dimension. $D_i$ for key, query, value dimension of each head, and $D^M$ for overall feature dimension.}
\label{table:detailed_setting}
\vspace{-1mm}
\end{table}

\vspace{-4mm}
\section{Dataset}

\begin{figure*}[h]
\begin{center}
\setlength{\tabcolsep}{0pt}
\begin{tabular}{@{}cc@{}}
\includegraphics[width=0.47\linewidth]{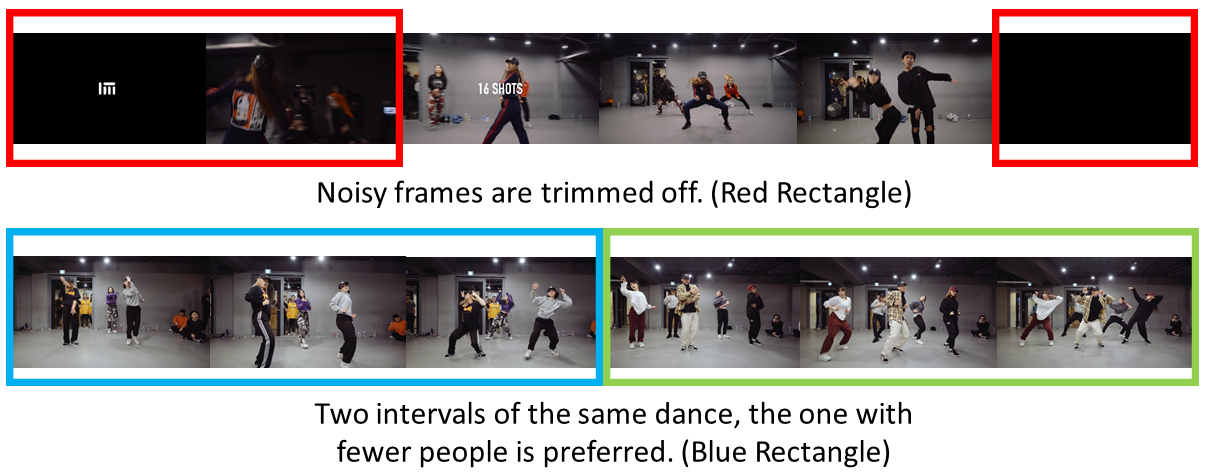} &
\includegraphics[width=0.47\linewidth]{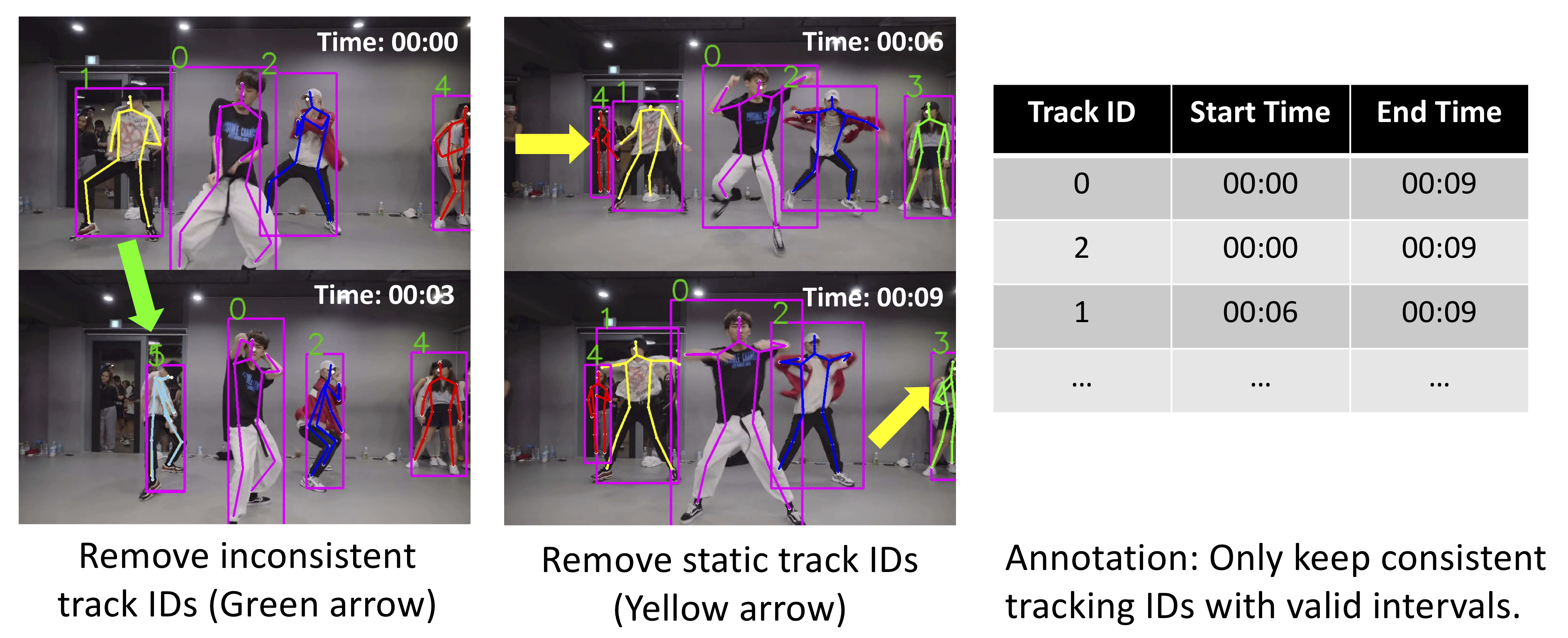}\\
(a) & (b)\\
\includegraphics[width=0.49\linewidth]{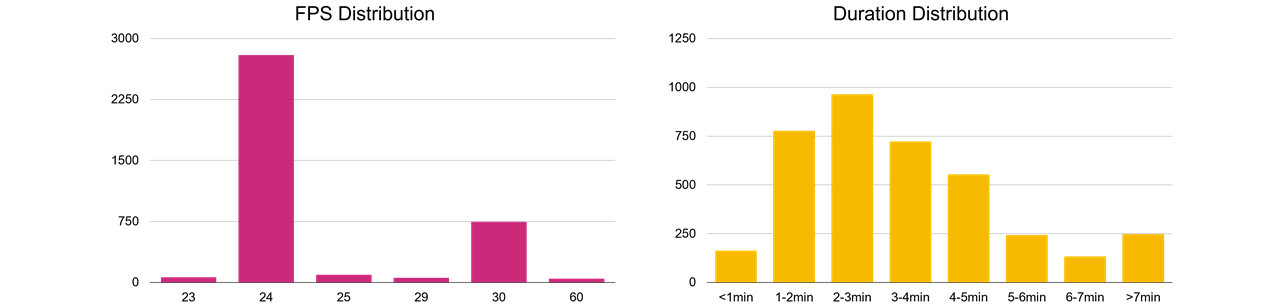} & 
\includegraphics[width=0.49\linewidth]{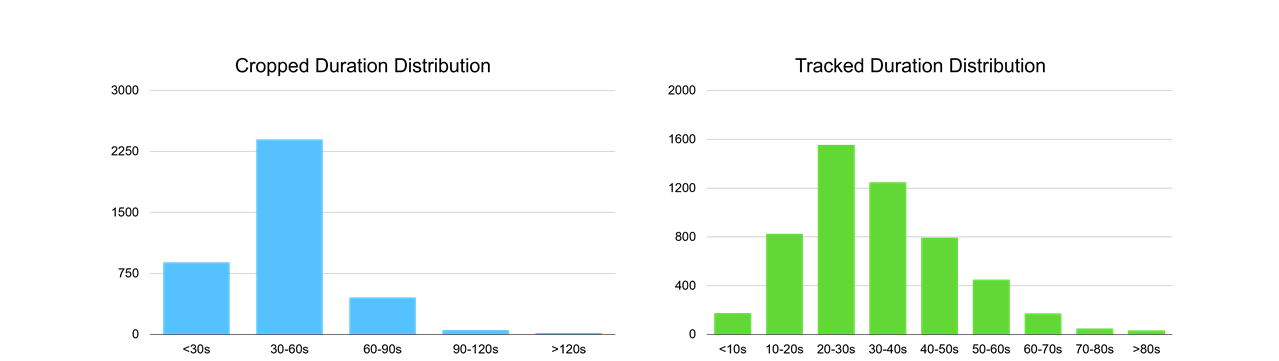}\\
(c) & (d)\\
\end{tabular}
\end{center}
\vspace{-2mm}
\caption{Our YouTube-Dance3D dataset. Figure (a) shows examples of video trimming. Figure (b) shows an example of valid pose tracking annotation. Figure (c) shows the distribution of original video FPS and duration. Figure (d) shows the duration distribution for cropped videos and tracked videos.}
\label{dataset_table}
\vspace{-1mm}
\end{figure*}

\begin{table*}[t!]
\begin{center}
\setlength{\tabcolsep}{4pt}
\begin{tabular}{@{}cc@{}} 
\setlength{\tabcolsep}{2pt}
\footnotesize{
\hspace{-4mm}
\begin{tabular}{@{}l||ccccccc@{}}
\hline

\\[-1pt]
Video Source & Video & Min. & Trim Seg. & Trim Min. & Track IDs & Track Min.\\[5pt]
\hline\hline
Urban Dance Camp & 197 & 400 & 193 & 173 & 9.1 & 201 \\[1pt] 
\hline
Mat  & 302 & 1865 & 462 & 241 & 8,7 & 248 \\[1pt] 
\hline
Movement Lifestyle & 478 & 1454 & 217 & 144 & 11.9 & 185\\[1pt] 
\hline
Snowglobe  & 1184 & 3568 & 873 & 654 & 10.2 & 727 \\[1pt] 
\hline
One Million Dance  & 1648 & 6301 & 2075 & 1494 & 14.5 & 1641 \\[1pt]
\hline\hline
Total  & 3809 & 13588 & 3820 & 2707 & 12.4 & 3002 \\[1pt]
\hline
\end{tabular}
}
& 
\footnotesize{
\setlength{\tabcolsep}{2pt}
\hspace{-5mm}
\begin{tabular}{@{}l||ccccccc@{}} 
 \hline
  & Source & Min. & Dancer & Audio & 3D & Public\\ \hline\hline
ChorRNN\cite{crnkovic2016generative}  & MoCap & 300 & 1 & \xmark & \cmark & \xmark \\ \hline
GrooveNet\cite{alemi2017groovenet}  & MoCap & 24 & 1 & \cmark & \cmark & \xmark \\ \hline
RobotKinect\cite{augello2017creative} & MoCap & - & 4 & \cmark & \cmark & \xmark \\ \hline
MelodyDance\cite{tang2018dance} & MoCap & 94 & - & \cmark & \cmark & \cmark \\ \hline
MikuDance\cite{yan2019convolutional} & Game & 600 & - & \cmark & \cmark & \xmark \\ \hline
YT2D\cite{1811.00818} & YouTube & 376 & - & \cmark & \xmark & \xmark \\ \hline\hline
DanceToMusic\cite{lee2019dancing2music} & YouTube & 4260 & - & \cmark & \xmark & \xmark \\ \hline\hline
Ours  & YouTube & 3002 & - & \cmark & \cmark & \cmark \\ \hline
\end{tabular}
}\\
\end{tabular}
\end{center}
\vspace{-1mm}
\caption{\small Left shows statistics of our dataset: the number of videos, duration in minutes, trimmed segments, duration after trimming, average number of track IDs per video, and total duration of all tracks. Right shows a comparison of our dataset to previous datasets.}
\label{table:ori-data-info}
\vspace{-4mm}
\end{table*}

We collected large amounts of high-quality videos from various dance channels on Youtube, and extracted 3D pose sequences with synchronized audios.
\vspace{-4mm}
\subsection{Dataset Collection}
We started by manually selecting popular YouTube dance studio channels and downloading both videos and tags. Our next-step data processing pipeline could be divided into four stages: {\textbf 1)} Quick annotation of dancing segments from untrimmed YouTube videos; {\textbf 2)} 2D pose detection and pose tracking; {\textbf 3)} Simple manual cleaning of correct tracking results; {\textbf 4)} 3D pose estimation from 2D and post-processing.
The first and third steps were introduced mainly because the Tubers edited the dance videos by inserting non-dance contents. If the video sources only contain clean dance clips, our pipeline would be fully automatic without manual annotation.

\vspace{-3mm}
\subsection{Video Statistics and Trimming}
We downloaded all the videos from five popular street dance channels and obtained 3809 videos in total. We filtered out the irrelevant contents e.g. dancer's daily life and trimmed the dance segments from the original videos. For each video, we annotated the start and end time of dance performance as shown in Figure~\ref{dataset_table}(a). The statistics of original and trimmed video segments are shown in Table~\ref{table:ori-data-info} and Figure~\ref{dataset_table}(c)(d). 

\vspace{-3mm}
\subsection{2D Pose Detection and Tracking}
We use YOLO-V3~\cite{redmon2018yolov3}, SimplePose~\cite{xiao2018simple} and LightTrack~\cite{ning2019lighttrack} to detect humans, estimate 2D pose, and track human pose sequences respectively.
Since the videos often contain multiple dancers and audiences, we kept top-5 largest human detection bounding boxes to reduce computation cost. 

\vspace{-3mm}
\subsection{Track Cleaning}
Since our collected online videos are completely unconstrained,
we observe two main issues: First, some tracks are audiences instead of dancers. Second, there are incorrect track id exchanges between dancers, which leads to the pose discontinuity. 
We performed manual annotations on pose sequences to reduce these types of noises. 
We ask volunteers to watch pose tracking visualization videos, and mark the correct start and end times for each track ids as shown in Figure~\ref{dataset_table}(b). We notice that imperfect tracking usually happens during group formation changes, where dancers occasionally occlude each other. 
However, this does not affect the overall data quantity and distribution because most correct tracks have sufficient durations. Figure~\ref{dataset_table}(d) shows the statistics after this data cleaning step.

\vspace{-3mm}
\subsection{3D Pose Estimation and Jitter Removal}
We applied VideoPose3d~\cite{pavllo:videopose3d:2018} to convert 2D pose sequences into 3D with 17 joints. 
Upon examining the results, we observed frequent motion jitters to be the main issue. Thus, we used Hodrick-Prescott (HP) filter~\cite{hodrick1997postwar} to remove these jitters.
HP filter separates a time-series $\mathbf{X}$=\{$x_t$\} into a trend component and a cyclical component, i.e. $x_t = x^{trd}_t + x^{cyc}_t$.
The components are determined by minimizing a quadratic loss function,
\begin{eqnarray}
\min_{\{ x^{trd}_{t}\}}\sum_{t}\left[x^{cyc}_t\right]^{2}+\lambda\left[x^{trd}_t-2x^{trd}_{t-1}+x^{trd}_{t-2}\right]^{2}
\end{eqnarray}

We applied this filter to the 3D coordinates of each joint separately, with $\lambda$=1 which empirically produces better result on our data.
All poses are bicubic interpolated into 24-fps to ensure consistent framerates.

\vspace{-3mm}
\subsection{Audio Processing}
Mel-Frequency Cepstral Coefficients (MFCC) are effective audio features widely used in various audio related tasks~\cite{shlizerman2018audio}.
We use LibROSA~\cite{mcfee2015librosa} to compute music features including MFCC and time intervals between beats.
The audio are standardized with 44.1Khz sample rate with EBU R128 loudness normalization via ffmpeg-normalize, after which features are extracted at 24-fps.

\vspace{-3mm}
\subsection{Dataset Analysis}
Compared to existing dance datatsets, our dataset has not only larger scale, but also higher quality and bigger variety.
As given In Table~\ref{table:ori-data-info}, the scale of YouTube-Dance3D exceeds the previous largest YT2D~\cite{1811.00818} by more than a magnitude order and comparable to a concurrent pose2D dataset introduced in ~\cite{lee2019dancing2music}. 
Moreover, our dataset possesses higher diversity due to the choice of urban dance, which by nature emphasizes choreography and variation. This is in contrast to other datasets that contain dances with obvious repetitive motion patterns e.g. Salsa, Tango, Ballet, and Waltz.  
Additionally, our data processing pipeline enables inexpensive data collection, making future expansion possible given new sources of dance video. 
We split data into approximately 80\% training and 20\% validation. For the convenience of model training, we divide data into segments of 20s, with 10s overlap between two consecutive segments. Each segment contains 480 frames. This results in 9561 training segments and 2136 validation segments. 

\vspace{-4mm}
\section{Experiments}
We first describe evaluation metrics including physical plausibility, beat consistency and generation diversity. Next, we show quantitative and qualitative results. 

\vspace{-4mm}
\subsection{Metrics}
Automatic evaluation metric has been a known challenge in many generative tasks. 
Recently~\cite{yan2019convolutional,lee2019dancing2music}, motion generation evaluation receives more attention, where FID-based metrics have been explored.
We share the similar insight on the aspect of evaluating generation diversity. Moreover, we further introduce to use a humanoid physics simulator to evaluate dance plausibility and a new beat consistency metric.

\vspace{-2mm}
\subsubsection{Physical Plausibility}
We measure the ratio of implausible frames that can not be executed by the humanoid inside Bullet simulator. 
Concretely, we measure two types of pose invalidity: 
1) \textit{Authenticity} is the ratio of frames where none of the joints exceeded its rotation limit. 
This ensures the pose is statically plausible, as can be performed by a normal human body.
2) \textit{Coherence} is the ratio of frames where the angular velocity of all joints stay within a realistic range. 
This ensures the motion between poses is dynamically plausible, preventing abnormal behaviour such as twitching.

\vspace{-2mm}
\subsubsection{Beat Consistency}
A good dancer knows to express their perception of beat by periodically changing their moves, 
in other words, to accompany musical beats with their motion beats.
We first extract motion beats from poses. Then we measure the similarities between generated motion beats and ground-truth motion beats.
We extract motion beats using the method of Kim~et~al.~\cite{kim2003rhythmic}, which detects zero-crossings of the joint angular acceleration. 
For two beats to match, we allow a flexibility of $2$ frames.
We compute \textit{precision}, \textit{recall}, and \textit{F-score} between two beat sequences.

\begin{figure}[t!]
\begin{center}   
\includegraphics[width=\linewidth]{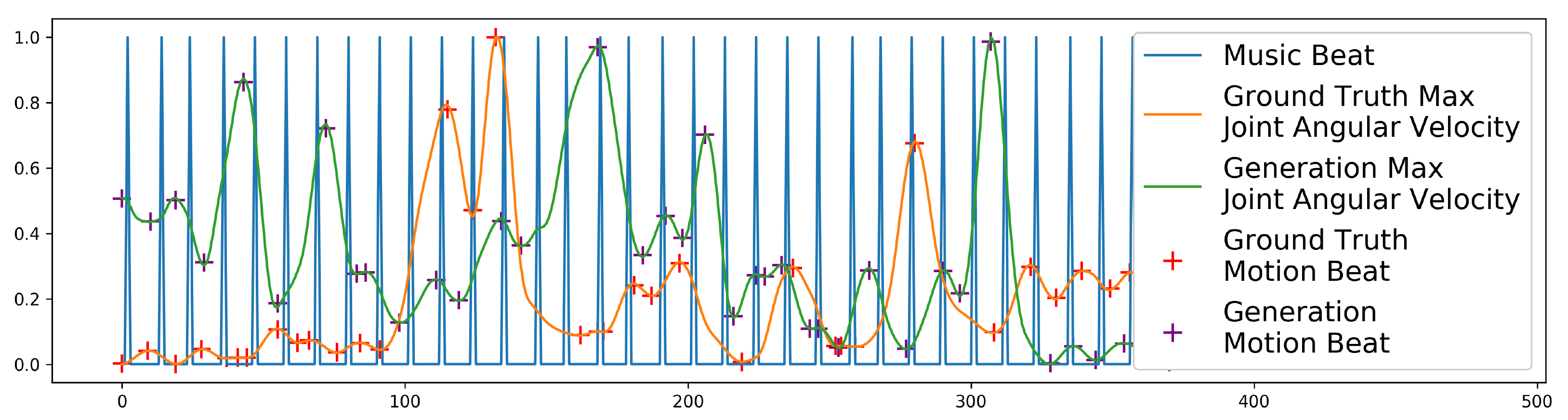}
\end{center}
\vspace{-2mm}
\caption{An example of detected music beats and motion beats.}
\label{motion_beat}
\vspace{-3mm}
\end{figure}

\vspace{-2mm}
\subsubsection{Diversity}
The complexity of choreography reflects in the composition of diverse body movements.
We measure the diversity aspect of generated dances via four aspects.
1) \textit{Fréchet Inception Distance (FID)} 
This refers to the default usage of FID~\cite{heusel2017gans}, measuring the difference between ground truth and generation feature distribution.
2) \textit{Inter-sequence Diversity (A-seq-D)}
We generate a large number of pose sequences, from which pairs of sequences are randomly selected. For each pair, we measure the L2 distance between their feature vectors. We use the average feature distance as the A-seq-D score.
3) \textit{Intra-sequence Diversity (I-seq-D)} 
Within a pose sequence, we divide it into chunks, and compute the feature distance among all possible pairs. This distance is averaged over all pairs from all sequences as the I-seq-D score.
4) \textit{Same-music Diversity (S-music-D)}
We generate multiple sequences given the same music, and compute the feature distances between these generations. We average this over all music as the S-music-D score. 

We obtain perceptual features of the dance with a dance style classifier. We first divide our dataset into $5$ categories based on the YouTube channel name, with balanced size between classes. Then we train a 2-block transformer with classification outputs, obtaining $61.0\%$ top-1 classification accuracy and $71.5\%$ top-2 classification accuracy.

\vspace{-4mm}
\subsection{Comparisons}
Previous work mainly focus on motion synthesis without audio input. Therefore, we first compare our model with them in non-audio setting. Then we compare variations of audio-enabled models.
\vspace{-2mm}
\subsubsection{acLSTM~\cite{li2017auto}}
acLSTM~\cite{li2017auto} is a widely-used model in dance motion synthesis from mocap data. 
It introduces an interval when ground-truth and model samples are used in the training process. This address the motion freeze issue of standard teacher-forcing training, which enables generating unlimited sequence length by learning from small amount of data. 
The model is a 3-layer LSTM with a hidden dimension of 1024. 
We follow the same settings as described in \cite{li2017auto} to train on our dataset. acLSTM is essentially a deterministic model, we evaluate its generative ability by feeding it additional information of different initial pose sequences with a length of 10.
\vspace{-2mm}
\subsubsection{ChorRNN~\cite{crnkovic2016generative2}}
ChorRNN~\cite{crnkovic2016generative2} is a mixture density model based on a 3-layer LSTM. In each time step, ChorRNN predicts a distribution of pose instead of deterministic coordinates. Since their code is not public, we re-implement their model and experiment with different number of mixtures.

\begin{table}[t!]
\begin{center}
\footnotesize{
\setlength{\tabcolsep}{1.5pt}
\begin{tabular}{@{}l||ccccccc@{}} 
 \hline
 & Coherence$\uparrow$ & Authenticity$\uparrow$  &  FID$\downarrow$  & A-seq-D$\uparrow$ & I-seq-D$\uparrow$\\ \hline\hline
Ground-Truth & 1 & 1 & 0 & 32.84 & 10.76 \\ \hline
acLSTM  & \textbf{0.9995} & \textbf{0.9998} & 3.94 & 12.99 & 3.55 \\ \hline
ChorRNN-5 & 0.83 & 0.75 & 2.56 & 30.48 & \textbf{21.26} \\ \hline
DLSTM & 0.94 & 0.88  & 2.47 & 29.07 & 12.06 \\ \hline
TSMT-noaudio & \textbf{0.97} & \textbf{0.96} & \textbf{0.53} & \textbf{32.52} & 7.98 \\ \hline
\end{tabular}
}
\end{center}
\vspace{-1mm}
\caption{\small Comparing different methods at the non-audio setting.}
\label{table:non-audio-res}
\vspace{-1mm}
\end{table}

\vspace{-3mm}
\subsection{Results}

\begin{figure*}[h]
\begin{center}
\begin{tabular}{c||c}
\hline
\rotatebox{90}{\scriptsize{acLSTM\cite{li2017auto}}} & \includegraphics[width=0.93\linewidth]{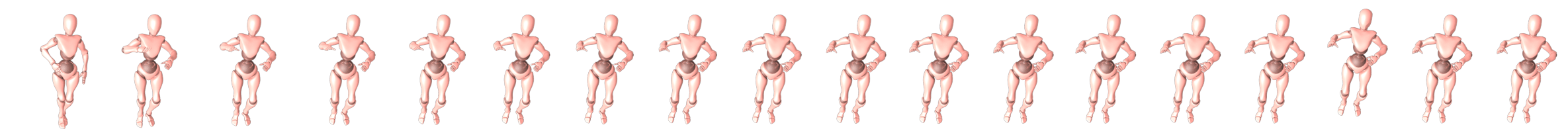}\\
\hline
\rotatebox{90}{\scriptsize{ChorRNN\cite{crnkovic2016generative}}} & \includegraphics[width=0.93\linewidth]{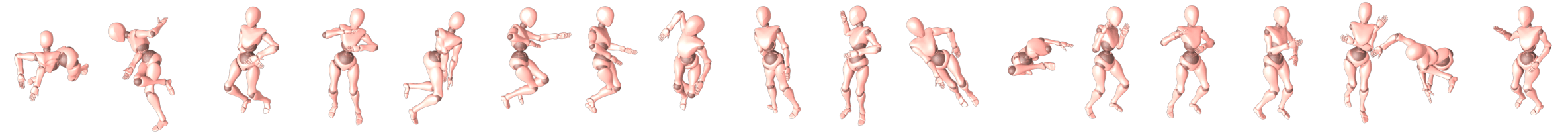}\\
\hline\hline
\multirow{3}{*}[-1.1em]{\rotatebox{90}{\scriptsize{Our TSMT-no-audio}}} & \includegraphics[width=0.93\linewidth]{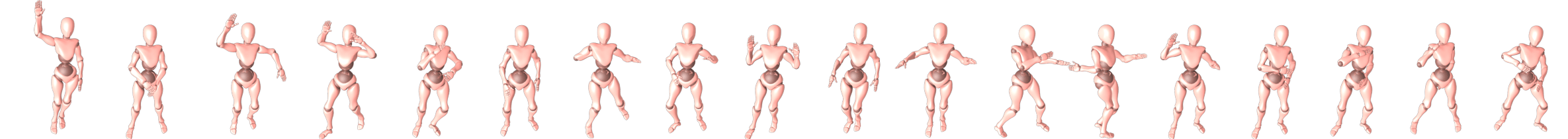}\\
\hhline{~|*1{-}|}
~ & \includegraphics[width=0.93\linewidth]{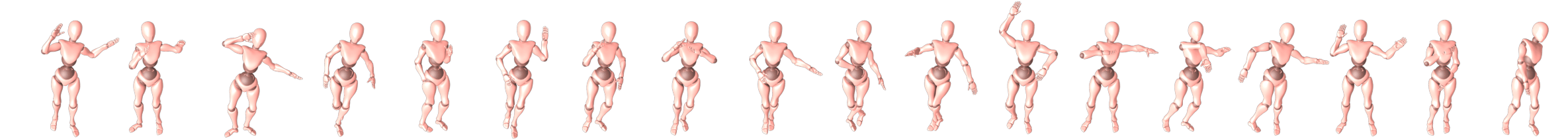}\\
\hhline{~|*1{-}|}
~ & \includegraphics[width=0.93\linewidth]{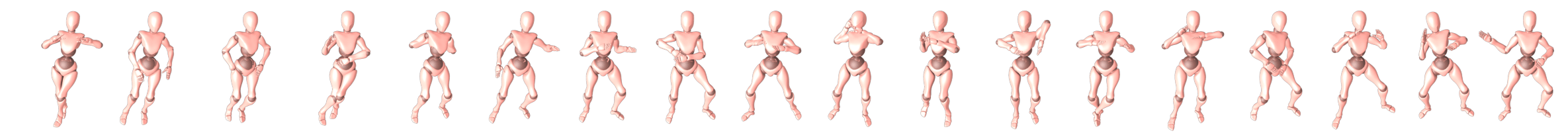}\\
\hline
\end{tabular}
\end{center}
\vspace{-1mm}
\caption{Qualitative comparison. Our model generates plausible, realistic, and diverse dances.}
\label{model_gen}
\end{figure*}

\subsubsection{Quantitative Results}
We first report experimental results with the non-audio setting in Table~\ref{table:non-audio-res}.
We generate for each model 1000 pose sequences for evaluation. All model use a random initial first pose to reduce the possible noise brought by generation in the first step. 
It can be seen that for acLSTM~\cite{li2017auto}, although it has the highest Coherence and Authenticity scores, it is unable to generate diverse dances as shown in the FID and Diversity scores. 
For ChorRNN~\cite{crnkovic2016generative}, although it has the highest intra-sequence diversity, many of its generations are hardly valid human poses. This can be seen from its low Coherence and Autheticity scores. We further experiment different number of mixtures which consists of its most important parameters. From the results in Table~\ref{table:mdn-res}, the high FID scores show that none of the settings can generate both realistic and diverse dances.
Regarding Coherence and Authenticity, our model scores comparably to acLSTM, while being able to generate more diverse motions close to the ground truth distribution.        
In the audio-enabled setting, We show baseline and ablation study results in Table~\ref{table:audio-res}. Single Transformer model refers to the baseline model that directly concatenates audio and pose data as input to a single-stream transformer. We also explore the effects of using different combinations of audio information for each model. 
We observe that all of our models have high Coherence and Authenticity scores. The beat-only models achieve higher diversity scores. This is because of less constraints imposed by the audio input.

\vspace{-2mm}
\subsubsection{Qualitative Results}
We show qualitative results for acLSTM, ChorRNN and our proposed TSMT-noaudio Model in Figure~\ref{model_gen}. All results are from a generation of 20 seconds at the non-audio setting, screen captured at the same time interval. It can be seen that acLSTM tends to quickly freeze to a generic static pose. ChorRNN generates invalid poses that can hardly be performed by any dancer. Our model is able to generate valid and diverse poses, which is consistent with the quantitative evaluation metric scores as reported in Table~\ref{table:non-audio-res}.

\vspace{-2mm}
\subsubsection{Computation Time}
We test the computation time of our transformer model and baselines models in the non-audio setting. We use a single GPU GTX 1080 to perform the running time evaluation. For training time, we report the average time per batch from our training log. For testing time, we generate 10 sequences with each model, each with 480 time steps. Results are shown in Table~\ref{table:training_time}. 
It can be seen that our transformer-based model outperforms LSTM-based model in terms of training efficiency, while remaining real-time (24-fps) in testing.

\begin{table}[t!]
\begin{center}
\footnotesize{
\setlength{\tabcolsep}{2pt}
\begin{tabular}{@{}l||ccccccc@{}} 
 \hline
 & Coherence$\uparrow$ & Authenticity$\uparrow$ & FID$\downarrow$ & A-seq-D$\uparrow$ & I-seq-D$\uparrow$ \\ \hline\hline
ChorRNN-1  & \textbf{0.94} & \textbf{0.94} & 8.42 & \textbf{40.39} & 17.32 \\ \hline
ChorRNN-5 & 0.83 & 0.75 & \textbf{2.56} & 30.48 & \textbf{21.26} \\ \hline
ChorRNN-10 & 0.75 & 0.54 & 17.30 & 28.35 & 17.99 \\ \hline
ChorRNN-20 & 0.74 & 0.55 & 17.29 & 24.88 & 16.66 \\ \hline
\end{tabular}
}
\end{center}
\vspace{-1mm}
\caption{\small Effect of mixture component number for ChorRNN~\cite{crnkovic2016generative}}
\label{table:mdn-res}
\vspace{-1mm}
\end{table}

\begin{table*}[t!]
\small
\begin{center}
\footnotesize{
\setlength{\tabcolsep}{9pt}
\begin{tabular}{@{}l||ccccccc@{}} 
 \hline
 & Coherence$\uparrow$ & Authenticity$\uparrow$ & Beat$\uparrow$ & FID$\downarrow$ & A-seq-D$\uparrow$ & I-seq-D$\uparrow$ & S-music-D$\uparrow$ \\ \hline\hline
Ground-Truth & 1 & 1 & 1 & 0 & 32.84 & 10.76 & 0 \\ \hline
Single Stream Transformer + Beat & 0.97 & 0.92 & 0.447 & 0.56 & \textbf{41.95} & \textbf{11.09} & 17.30 \\ \hline
Single Stream Transformer + MFCC & 0.96 & 0.92 & 0.445 & 0.43 & 38.69 & 9.69 & 16.96 \\ \hline
Single Stream Transformer + Beat + MFCC & 0.97 & 0.93 & 0.439 & 1.27 & 38.86 & 10.88 & 16.00 \\ \hline
Two Stream Transformer + Beat & 0.97 & 0.93 & \textbf{0.451} & 1.43 & 40.90 & 9.49 & \textbf{18.11} \\ \hline
Two Stream Transformer + MFCC & 0.96 & 0.92 & 0.430 & 1.46 & 33.37 & 9.41 & 16.07\\ \hline
Two Stream Transformer + Beat + MFCC & 0.97 & 0.93 & 0.449 & \textbf{0.21} & 36.44 & 10.02 & 16.16 \\ \hline
\end{tabular}
}
\end{center}
\vspace{-1mm}
\caption{\small Comparing different methods at the audio-enabled settings.}
\label{table:audio-res}
\vspace{-1mm}
\end{table*}

\begin{table}[t!]
\begin{center}
\footnotesize{
\setlength{\tabcolsep}{3pt}
\begin{tabular}{@{}l||ccccccc@{}} 
 \hline
~ & acLSTM & ChorRNN & DLSTM & Our TSMT\\
\hline\hline
Train per Batch & 2.40s & 0.31s & 0.39s & \textbf{0.23s}\\
\hline
Test per Step &\textbf{0.55ms} & 62.5ms & 15.7ms & 19.4ms \\
\hline
\end{tabular}
}
\end{center}
\vspace{-1mm}
\caption{\small Training and testing time comparison.}
\label{table:training_time}
\vspace{-1mm}
\end{table}

\vspace{-4mm}
\subsection{Human Evaluation}
We conduct human evaluation to verify whether our proposed automatic metrics are consistent with human judgement. We make use of Amazon Mechanical Turk (AMT) and pay workers to perform a crowd-sourced evaluation. We restrict to US-based workers who have at least 85\% acceptance score.

\vspace{-2mm}
\subsubsection{Physical Plausibility and Beat} 
We first obtain Authenticity, Coherence, and Beat scores computed with our automatic metrics. Then for each metric, we divide the score into three levels of high, middle and low.  
We randomly sample 60 pairs from three different level combinations, namely high-mid, high-low, and mid-low. Then we resort to workers and ask them to select the better one from each pair. 
Each test contains 20 evaluation pairs and 3 validation pairs. The validation pairs contain a ground-truth and an artificially noised motion sequence. We use this as a hidden test to filter out the workers who are inattentive or intentionally abusing. We take the answer from AMT workers via majority-voting, compare with known answer decided by actual scores in each pair and calculate the consistency between two answers. 
We observe that Authenticity and Coherence have high consistency with human evaluations which verify the effectiveness of our physical plausibility metrics. However, the consistency for beats is relatively low for High-Mid and Mid-Low tests, while High-Low test has higher consistency. We believe that the average person does not have high sensitivity for beats as machine measurements do. In other words, some of AMT workers might be incapable of distinguishing whether the dance follows the beat or not when the differences are small. Further studies are needed in the future for deeper beat analysis.

\vspace{-2mm}
\subsubsection{Overall Quality}

\begin{figure}[t!]
\vspace{-2mm}
\begin{center}   \includegraphics[width=\linewidth]{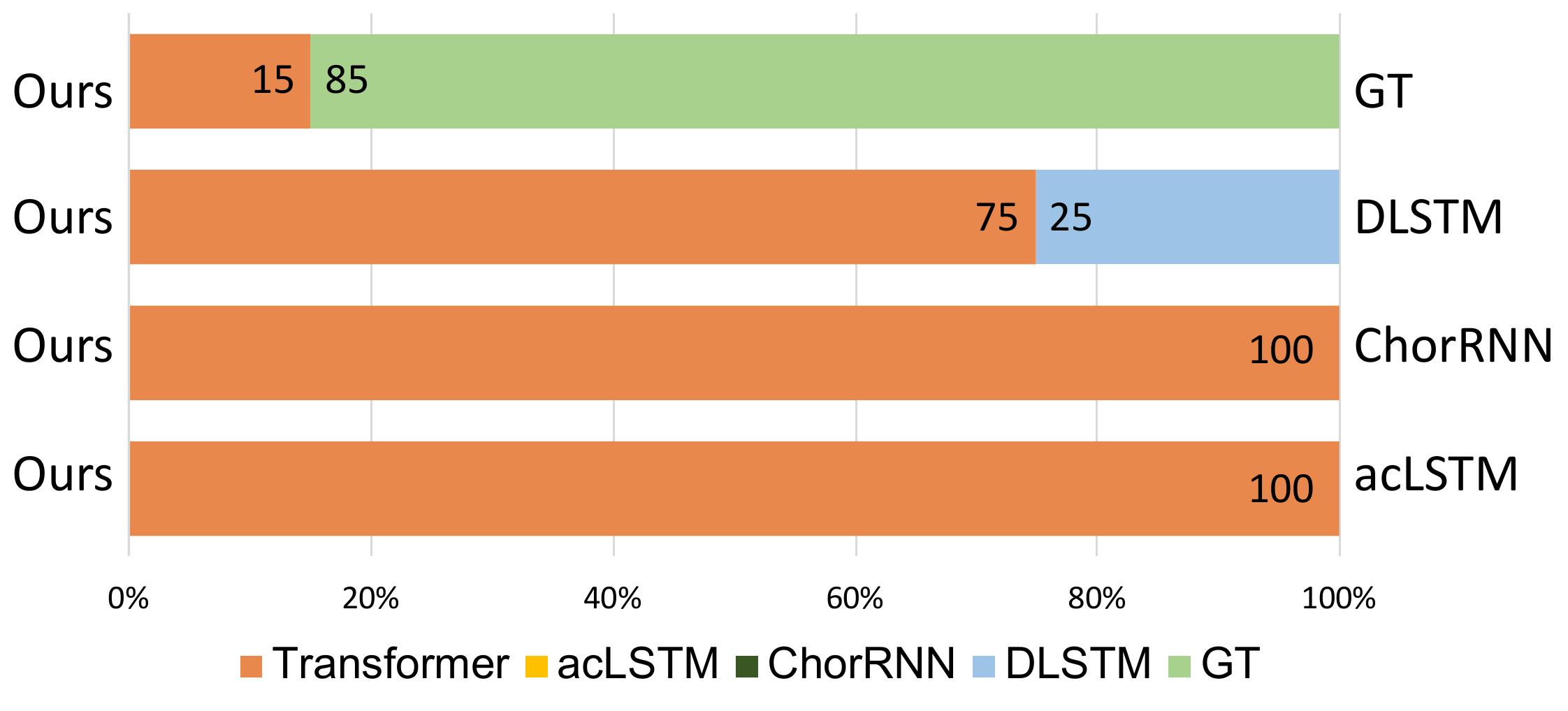}
\end{center}
\vspace{-3mm}
\caption{AMT user study on overall generation quality.}
\label{quality_human_eval}
\vspace{-5mm}
\end{figure}

We randomly select 200 pairs generated by acLSTM, ChorRNN, Discrete-LSTM, and our TSMT model (non-audio setting) in different combinations, and ask workers to pick the more preferable one with better quality. A comparison between our result and ground-truth is also included. Similarly, we use validation questions to filter out noisy annotations. Figure~\ref{quality_human_eval} shows the pair-wise comparison results. It can been seen that our model is obviously better than acLSTM and ChorRNN baselines, also superior than LSTM with discrete representation. However, there is still a gap between our synthetic motions and the ground truth, and the main reason is that our synthetic motion sequences are based on sampling in each timestep, there is a chance that low probability pose gets sampled which introduce noise to the sequence generation. It is possible to eliminate this type of noise by applying some constraints during pose sampling. 


\begin{table}[t!]
\small
\begin{center}
\footnotesize{
\setlength{\tabcolsep}{10pt}
\begin{tabular}{@{}l||ccccccc@{}} 
 \hline
 & High-Low & High-Mid & Mid-Low & Total  \\ \hline\hline
Authenticity  & 0.9 & 1.0 & 0.65 & 0.85  \\ \hline
Coherence & 0.95 & 0.95 & 0.9 & 0.933  \\ \hline
Beat & 0.65 & 0.55 & 0.4 & 0.533 \\ \hline
\end{tabular}
}
\end{center}
\vspace{-1mm}
\caption{\small AMT user study result to evaluate the consistency between automatic metrics and human judgement.}
\label{table:bullet_beat_human_eval}
\vspace{-1mm}
\end{table}

\vspace{-4mm}
\section{Discussion and Conclusion}
We proposed a complete system for dance motion synthesis from audio input, and we have shown that we can handle highly diverse dance movements using widely available online dance videos as training. 
We have also introduced a new large-scale motion dataset, as well as new evaluation metrics in terms of quality, diversity and musicality. Our proposed conditional generative model also outperformed existing methods.
We also conducted a study that indicates the effectiveness of our model and proposed metrics.

\vspace{-4mm}
\subsection{Limitation and Future Work}
Since our data is collected from videos, the number of joints depends on 3D pose estimation method which usually does not take finger animations into account. An interesting future direction is to extract finger joints and facial expressions as well from videos so that we are able to produce expressive dance data. Futhermore, we are interested in using motion capture data to denoise and improve the quality of 3D pose sequence extracted from online videos. 

We have used audio representations such as MFCC features and beat, but according to professional dancers and choreographers, they tend to also follow additional musical layers including bass, lyrics, etc. The exploration for more complex audio features is therefore particularly intriguing, as well as the analysis of interactions between dancers and crowds.



\vspace{-4mm}
\section{Acknowledgements}
This research was funded by in part by the ONR YIP grant N00014-17-S-FO14, the CONIX Research Center, a Semiconductor Research Corporation (SRC) program sponsored by DARPA, the Andrew and Erna Viterbi Early Career Chair, the U.S. Army Research Laboratory (ARL) under contract number W911NF-14-D-0005, Adobe, and Sony.
We thank Yajie Zhao, Mingming He for prof reading, Zhengfei Kuang for editing the demo, Marcel Ramos and Kalle Bladin for the character rigging and rendering, and Pengda Xiang for the data annotation. 

\bibliographystyle{eg-alpha-doi} 
\bibliography{egbibsample}       


\newpage


\end{document}